\documentclass[sigconf]{acmart}

\usepackage{subcaption}
\usepackage{algorithm}
\usepackage{algpseudocode}
\usepackage{graphicx}
\usepackage{multirow}
\usepackage{booktabs}
\usepackage[table]{xcolor}
\usepackage{balance}

\AtBeginDocument{%
  }

\copyrightyear{2025}
\acmYear{2025}
\setcopyright{cc}
\setcctype{by-nc}
\acmConference[CIKM '25]{Proceedings of the 34th ACM International Conference on Information and Knowledge Management}{November 10-14, 2025}{Seoul, Republic of Korea}
\acmBooktitle{Proceedings of the 34th ACM International Conference on Information and Knowledge Management (CIKM '25), November 10--14, 2025, Seoul, Republic of Korea}\acmDOI{10.1145/3746252.3761260}
\acmISBN{979-8-4007-2040-6/2025/11}

\begin{document}

\title{Seeing Through the Blur: Unlocking Defocus Maps for Deepfake Detection}


\author{Minsun Jeon}
\orcid{0009-0005-8519-4979}
\affiliation{%
  \institution{Sungkyunkwan University}
  \department{Computer Science \& Engineering Dept.}
  \city{Suwon}
  \country{Republic of Korea}}
\email{minsun9602@g.skku.edu}

\author{Simon S. Woo}
\orcid{0000-0002-8983-1542}
\authornote{Corresponding author}
\affiliation{%
  \institution{Sungkyunkwan University}
  \department{Computer Science \& Engineering Dept.}
  \city{Suwon}
  \country{Republic of Korea}}
\email{swoo@g.skku.edu}


\begin{abstract}
The rapid advancement of generative AI has enabled the mass production of photorealistic synthetic images, blurring the boundary between authentic and fabricated visual content. This challenge is particularly evident in deepfake scenarios involving facial manipulation, but also extends to broader AI-generated content (AIGC) cases involving fully synthesized scenes. As such content becomes increasingly difficult to distinguish from reality, the integrity of visual media is under threat.
To address this issue, we propose a physically interpretable deepfake detection framework and demonstrate that defocus blur can serve as an effective forensic signal. Defocus blur is a depth-dependent optical phenomenon that naturally occurs in camera-captured images due to lens focus and scene geometry. In contrast, synthetic images often lack realistic depth-of-field (DoF) characteristics.
To capture these discrepancies, we construct a defocus blur map and use it as a discriminative feature for detecting manipulated content. Unlike RGB textures or frequency-domain signals, defocus blur arises universally from optical imaging principles and encodes physical scene structure. This makes it a robust and generalizable forensic cue. Our approach is supported by three in-depth feature analyses, and experimental results confirm that defocus blur provides a reliable and interpretable cue for identifying synthetic images. We aim for our defocus-based detection pipeline and interpretability tools to contribute meaningfully to ongoing research in media forensics. The implementation is publicly available at: \url{https://github.com/irissun9602/Defocus-Deepfake-Detection}
\end{abstract}




\maketitle

\section{Introduction}
\begin{figure}[t]
    \centering
    \includegraphics[width=1\linewidth]{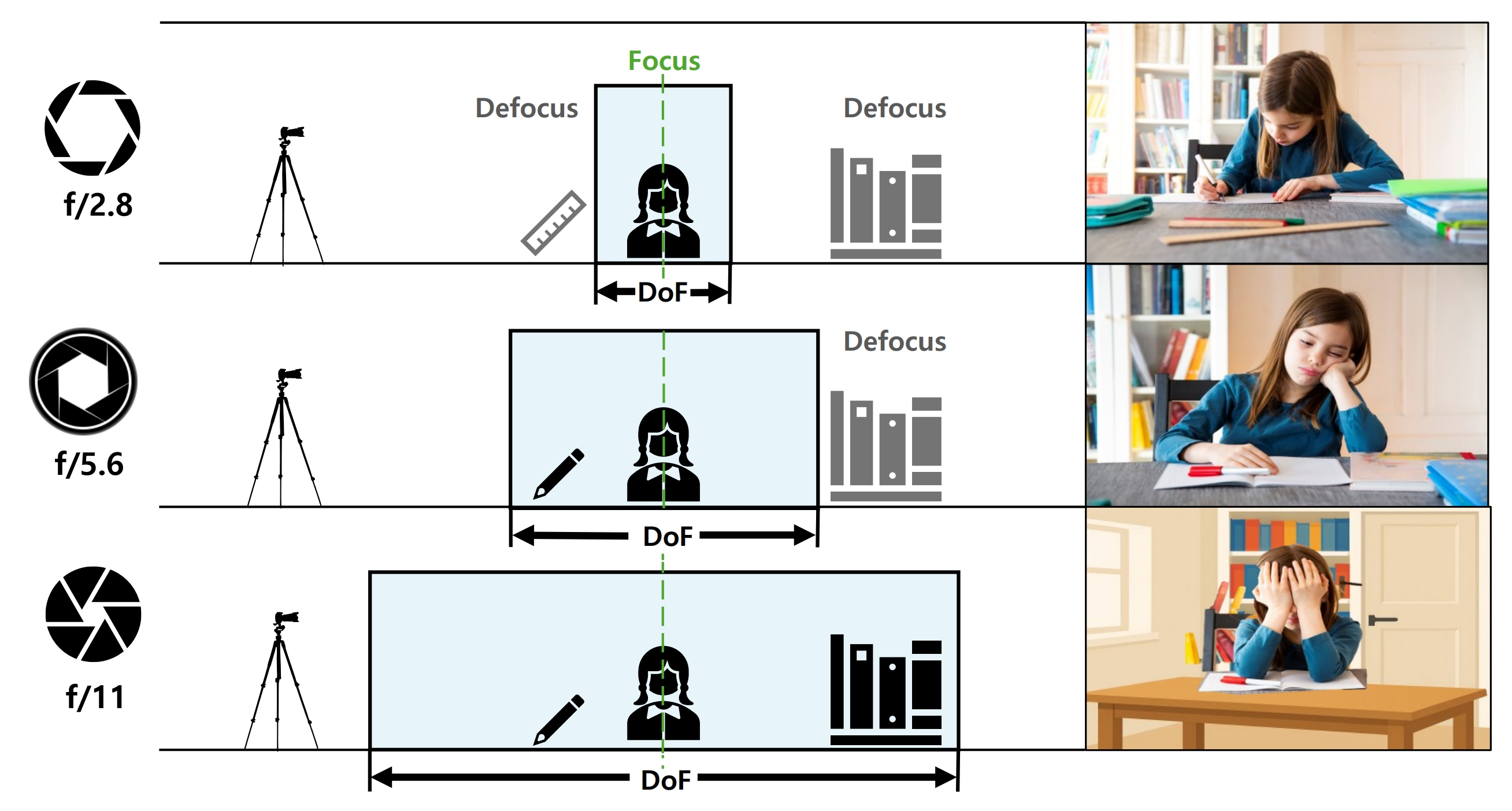}
    \caption{Defocus induced by Depth of Field (DoF) as a forensic cue. This figure illustrates how Depth of Field naturally produces defocus, forming the physical basis of our defocus blur analysis for deepfake detection. Here, $f$ denotes the aperture value (f-number). With a wide aperture ($f/2.8$), only the focal subject appears sharp. A moderate aperture ($f/5.6$) expands DoF, adding nearby objects into focus. At a narrow aperture ($f/11$), both foreground and background remain sharp. These optically induced defocus patterns are leveraged to construct Defocus Blur Maps as a forensic cue.
    }
    \Description{Diagram showing how depth of field affects sharpness at three aperture settings: f/2.8 produces sharp foreground and blurred background, f/5.6 increases sharpness for nearby objects, and f/11 keeps both foreground and background in focus.}
    \label{fig:dof}
\end{figure}

As synthetic media becomes increasingly diverse and sophisticated, it is critical to explore whether new modalities of image representation can provide universal and robust signals for deepfake detection. While deep learning-based detection models have demonstrated impressive success~\cite{1,2,3,4,5,6,7,8,9,10,11,12,13,14,15,16,17,18,19,20,21,22,23,24,25,26,27,28,29,30,31,32,33,34,35,36,37,38,39,40,41}, they often rely on RGB textures or frequency-domain signals that are susceptible to adversarial manipulations and lack interpretability. This black-box nature makes it challenging to understand why certain decisions are made, limiting the trustworthiness of detection results and impeding generalization to unseen types of forgeries. This limitation underscores the need for physically grounded features. In this work, we investigate optical defocus blur as such a modality, leveraging its inherent connection to scene geometry and camera optics to distinguish real from synthetic imagery.

To provide such a physically grounded cue, we focus on optical defocus blur, which arises naturally in real images due to lens focus and scene geometry. Depth of field (DoF) governs how objects at varying distances from the focal plane appear sharp or blurred, creating predictable defocus patterns in authentic photographs. Figure \ref{fig:dof} illustrates how the distance of objects from the camera affects their sharpness and blur, as well as how depth of field changes with different aperture settings also known as f-stop values. However, synthetic images generated by GAN~\cite{goodfellow2020generative}, VAE~\cite{kingma2013auto}, or diffusion model~\cite{ho2020denoising} are produced through algorithmic pipelines that typically do not explicitly model optical constraints. As a result, these images often exhibit globally sharp or physically inconsistent blur patterns, lacking the depth-dependent blur that naturally emerges in real scenes. This limitation is consistent with prior findings in blur synthesis research, which show that synthetic blur often fails to capture the complex characteristics of real optical defocus~\cite{rim2022realisticblursynthesislearning,zhang2020deblurringrealisticblurring}. In particular, generative models tend to produce all-in-focus images or ignore physically plausible defocus effects, prioritizing perceptual quality over optical realism. For example, VAE-based models are known to yield overly smooth or blurry outputs due to their reconstruction loss, prompting the use of anti-blur loss functions~\cite{bredell2023explicitly}. GAN-based models, on the other hand, often adopt perceptual losses~\cite{dosovitskiy2016generating} or multi-view fusion strategies~\cite{luo2022point} to enhance sharpness. While these techniques improve visual appeal, they are not designed to reproduce the depth-dependent blur patterns governed by scene geometry and lens optics~\cite{naderi2020generating, ali2024upscaling}. 

Although defocus blur has been widely studied for depth estimation, its use as a discriminative feature for detecting manipulated images has been rarely explored. 
This gap highlights the novelty of our approach, positioning defocus blur as a physically grounded signal that generative models inherently struggle to replicate. In addition, defocus blur enhances explainability in forensic analysis. As a perceptually interpretable optical cue, defocus enables both experts and non-experts to intuitively understand the basis of model decisions. This contrasts with previous methods that are difficult to visualize or explain, thereby positioning defocus as a trustworthy signal for transparent and human-understandable deepfake detection.

To guide our investigation and assess the proposed method, we formulate the following research questions:
\begin{itemize}
    \item \textbf{RQ1}: Do Defocus blur patterns differ between real and manipulated images?
    \item \textbf{RQ2}: Can these features improve the performance of deepfake detection when integrated into a detection pipeline?
    \item \textbf{RQ3}: Does the trained model actively utilize defocus-based cues in its decision-making process?
\end{itemize}

Our main contributions, corresponding to the above research questions, are as follows:
\begin{itemize}
    \item \textbf{Defocus blur for deepfake detection (RQ2)}: We propose a defocus-based detection method that distinguishes real and synthetic images by leveraging depth-dependent blur differences.
    \item \textbf{In-depth analysis for feature effectiveness (RQ1)}: We conduct three analyses, providing visual evidence and statistical validation that real and fake images exhibit distinct defocus blur characteristics.
    \item \textbf{Attribution analysis demonstrating how defocus cues are utilized (RQ3)}: We validate model interpretability through SHAP-based attribution, showing that defocus-aware models actively use blur cues for robust and transparent detection.
\end{itemize}

\section{Related Work}
\subsection{Deepfake Detection}
\textbf{Deep Learning-based Methods.}
Spatial-based methods such as XceptionNet~\cite{rossler2019faceforensics++} and MesoNet~\cite{afchar2018mesonet} detect manipulations using pixel-level features in the RGB domain but often struggle against advanced generative models. Frequency-based methods, including F3Net~\cite{qian2020thinking}, analyze local frequency artifacts but typically lose spatial precision due to domain transformation. To overcome these limitations, hybrid approaches like SPSL~\cite{liu2021spatial} focus on local texture-level artifacts arising from upsampling. 

\noindent \textbf{Physically Grounded Methods.}
In pursuit of more interpretable features, researchers have turned their attention to physically groun-ded cues. 
Depth-guided approaches, for example, leverage monocular depth estimation to extract geometric structure for forgery detection. 
Liang et al.~\cite{liang2023depth} proposed a depth-map-guided triplet network for facial forgeries. While effective in face-centric settings, such methods are computationally heavy and often unreliable under occlusion or complex scenes. 
Other studies have explored sensor-level fingerprints such as PRNU-based detection~\cite{prnu}, which capture device-specific noise patterns but degrade under compression or camera changes. 

\noindent \textbf{Handcrafted Feature-based Methods.}
Traditional handcrafted methods relied on interpretable cues such as SURF descriptors~\cite{zhang2017automated}, head pose~\cite{yang2019exposing}, or biological signals~\cite{ciftci2020fakecatcher}, but were often limited to facial content. In contrast, our method leverages defocus blur—a universal, depth-related optical cue present in all camera-captured images. As a content-agnostic signal, it extends the applicability of handcrafted detection to broader media contexts.

\subsection{Defocus Map Estimation}
A defocus map quantifies the level of defocus blur or the size of the circle of confusion (CoC) at each pixel in a blurred image. Estimating defocus blur from images has been approached in various ways and can be broadly categorized into three groups: \textbf{edge-based}~\cite{elder1998local, zhuo2011defocus}, \textbf{region-based}~\cite{shi2015just, trouve2011single}, and \textbf{CNN-based}~\cite{lee2019deep, ma2021defocus} methods. Edge-based methods typically analyze gradient sharpness near edges, while region-based approaches examine texture or intensity variation within local patches. More recent CNN-based methods leverage small datasets to learn complex blur representations directly from data. These defocus estimation techniques have been extensively studied because defocus blur encodes valuable physical information about the scene and camera optics. Specifically, defocus has been used to estimate depth maps~\cite{ding2017robust, wijayasingha2024camera}, restore sharp images~\cite{ma2021defocus}, assist in auto-focus systems~\cite{ding2017robust}, and support scene understanding tasks in fields such as robotic vision~\cite{cristofalo2017out}. In these applications, the spatial variation of defocus blur serves as a low-level geometric cue that reflects how objects are positioned relative to the focal plane.

Despite its established utility in physically grounded image analysis, defocus blur has rarely been explored as a discriminative feature for detecting manipulated or synthetic content. This work proposes a novel perspective by leveraging defocus blur as a primary signal for deepfake detection, aiming to identify inconsistencies in physical blur patterns that are often overlooked by existing methods.

\section{Proposed Method} \label{sec:method}
\subsection{Defocus Blur-based Detection Approach}
\subsubsection{\textbf{Defocus Blur as a Discriminative Feature}}
The amount of blur is governed by several physical parameters, including lens configuration, aperture size, and object distance.
We exploit defocus blur as a discriminative cue by explicitly estimating a defocus map for each input image, capturing pixel-wise blur intensity that reflects depth-dependent optical effects. These maps are then fed into a detection model trained to distinguish real and fake content. By exploiting this gap, our method detects manipulations by identifying inconsistencies in spatial blur distribution that deviate from real-world imaging characteristics.

\subsubsection{\textbf{Defocus Blur Map Estimation}}
\label{sec:defocus_generator}
A defocus blur map is a per-pixel representation that quantifies the degree of defocus at each point in an image. It reflects how far each pixel deviates from optical sharpness, which occurs when a region lies outside the camera’s focal plane. 

To estimate this blur, we adopt an edge-based algorithm that infers defocus from the degradation of gradient strength along with image contours. Unlike depth-based approaches that rely on geometric priors or stereo input, edge-based methods offer a lightweight and content-agnostic solution by analyzing local sharpness variation. This aligns with the concept of Depth of Field (DoF), where sharp edges typically lie in the focal region, and blurred edges indicate out-of-focus areas.

Specifically, we implement the method of Zhuo and Sim~\cite{zhuo2011defocus}, which computes the ratio of gradient magnitudes at two Gaussian scales, \(\sigma_1\) and \(\sigma_2\). The final defocus estimation is given by:
\begin{equation}
\sigma = \sqrt{ \max\left( \frac{R^2 \cdot \sigma_1^2 - \sigma_2^2}{1 - R^2 + \epsilon}, \, 0 \right) },
\label{eq:defocus}
\end{equation}
where \(R\) denotes the gradient magnitude ratio, and \(\epsilon\) is a small constant for numerical stability. The implementation was adapted from a publicly available MATLAB version~\cite{github_matlab_defocus}, and modified to a generalized formulation. Specifically, we replaced the original formatting function with a fast guided filter for improved edge-preserving smoothing. The upper part of Figure~\ref{fig:overview} illustrates the overall defocus blur map estimation process. To quantify the computational overhead of this module, we measured runtime and memory usage on a single GPU with a batch size of 64, excluding the first 5 warm-up batches and averaging over the next 30 batches.
As shown in Table~\ref{tab:defocus_overhead}, the edge map computation dominates the runtime (~582 ms), while other stages contribute negligibly (<7 ms). In terms of memory, the closed-form matting step reached the highest peak usage (~848 MB).
\begin{table}[t]
\centering
\caption{Runtime and memory overhead of the defocus estimation module (batch size = 64). We exclude the first 5 warm-up batches and report the mean over the next 30 batches.The Edge Map stage dominates runtime, while other steps incur relatively minor overhead, indicating that the additional defocus estimation module is lightweight and practical for integration.}
\label{tab:defocus_overhead}
\resizebox{\columnwidth}{!}{
\begin{tabular}{lccc}
\toprule
\textbf{Stage} & \textbf{Avg. Time (ms)} & \textbf{Peak Mem. (MB)} \\
\midrule
RGB to Gray             & 0.221   & 475.6 \\
Edge Map                & 582.088 & 497.4 \\
Sparse Blur Map         & 6.475   & 781.9 \\
Closed-Form Matting     & 6.394   & 847.9 \\
\midrule
\textbf{Total}          & 595.178 & --  \\
\bottomrule
\end{tabular}
}
\end{table}

\subsubsection{\textbf{Defocus-based Forgery Detection Pipeline}}
\begin{figure}[t]
    \centering
    \includegraphics[width=1\linewidth]{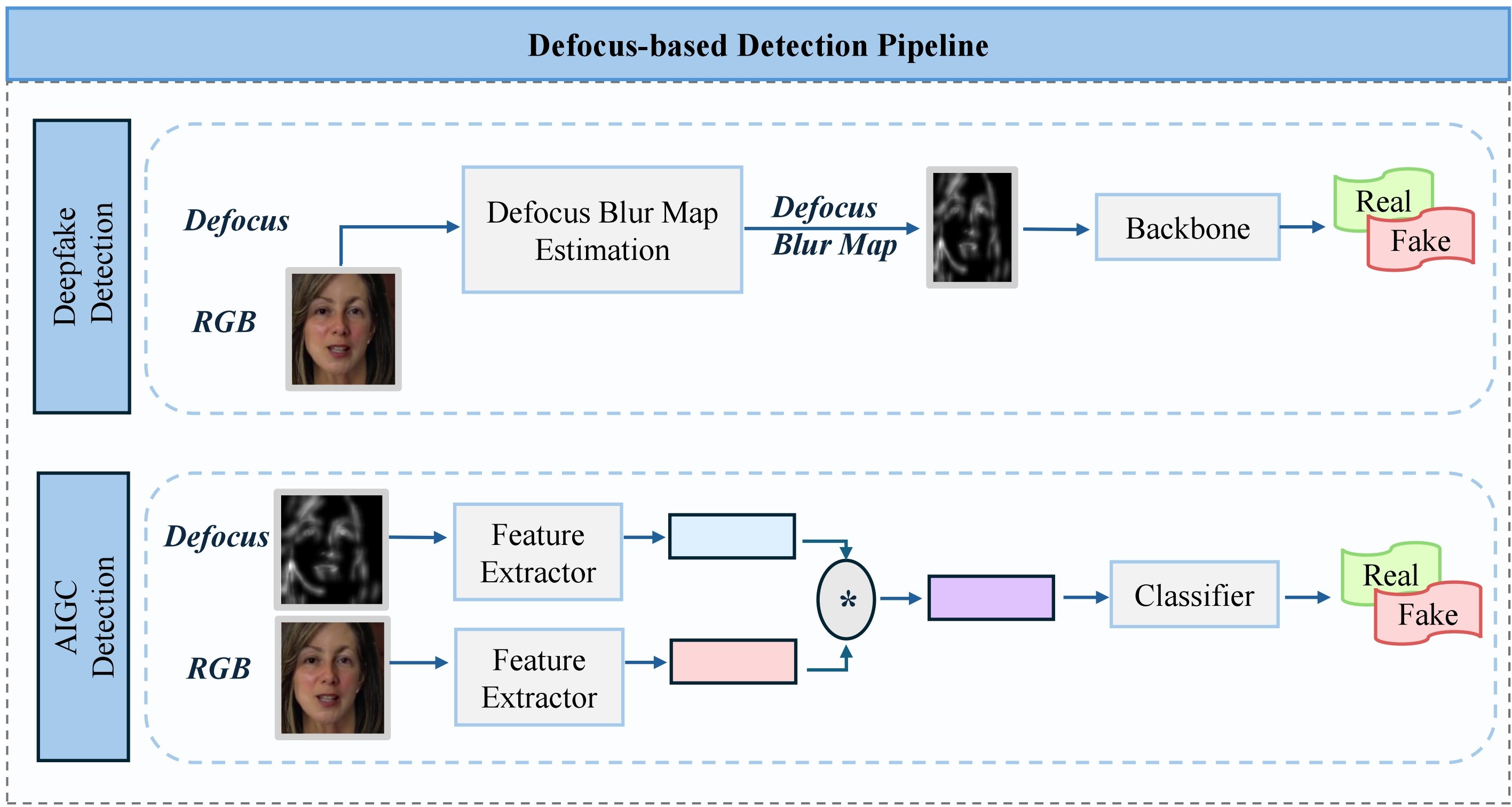}
    \caption{
    Overview of the proposed defocus-based forgery detection pipeline. The framework is implemented in two configurations: (1) a defocus-only deepfake detection pipeline, where defocus blur maps serve as the sole input to a classifier, and (2) a dual-branch RGB+Defocus fusion pipeline for AIGC detection, which combines defocus and RGB features to improve classification.}
    \Description{
    Two detection pipelines are shown. The upper pipeline uses only defocus maps for deepfake detection. 
    The lower pipeline fuses RGB and defocus features for AIGC detection using dual-branch architecture.
    }
    \label{fig:pipeline}
\end{figure}

As illustrated in Figure~\ref{fig:pipeline}, our proposed pipeline adopts two configurations depending on the task domain.

For deepfake detection, we adopt a defocus-only approach, where the input consists solely of defocus blur maps derived from RGB images. These maps are estimated using our DefocusMapGenerator (Section~\ref{sec:defocus_generator}) and are fed into a full XceptionNet classifier without any RGB input. This design allows us to assess the standalone effectiveness of defocus blur as a physically grounded feature for detecting facial manipulations. The upper part of Figure~\ref{fig:pipeline} visualizes this single-modality architecture, evaluated on the FaceForensics++~\cite{rossler2019faceforensics++} dataset.

For AIGC detection, where both foreground and background content may be synthetically generated, we employ a dual-branch fusion model. RGB and defocus inputs are processed separately through two parallel XceptionNet encoders with matching architectures but different input channels. Their resulting feature maps are fused via hadamard product and passed to a classifier head. This fusion allows defocus information to modulate RGB-based features as an auxiliary physical cue, enhancing detection in complex synthetic scenes. The lower part of Figure~\ref{fig:pipeline} illustrates this multimodal configuration, used in experiments with the SYNDOF~\cite{lee2019deep} dataset.

\subsection{Defocus Blur Map Feature-level Analysis}
\begin{figure*}[t]
    \centering
    \includegraphics[width=1\linewidth]{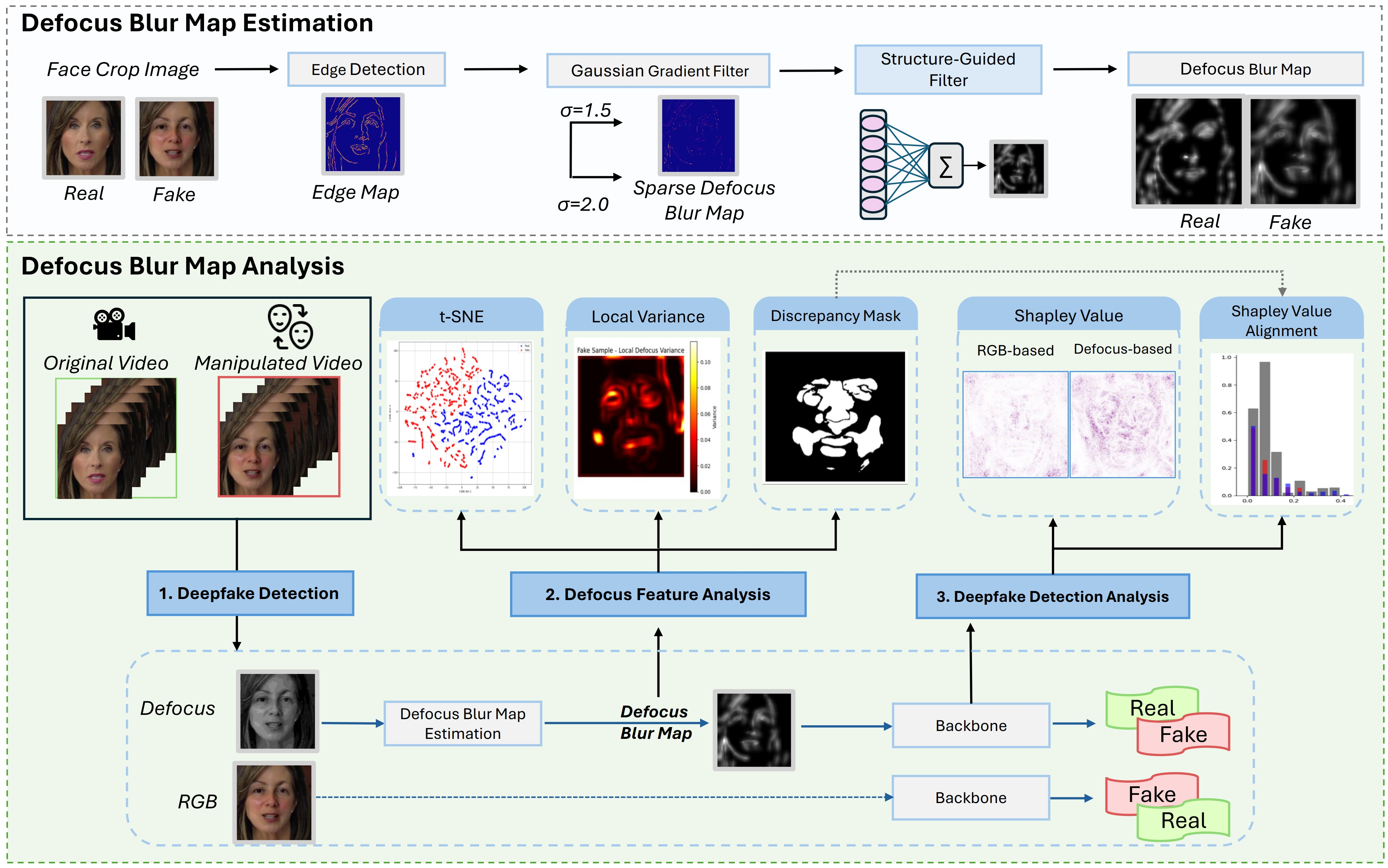}
    \caption{
    Overview of the Defocus blur analysis framework. The framework is designed to validate defocus blur as a forensic signal. It consists of a defocus blur map estimation module and two analysis parts: (1) feature-level analysis (t-SNE, local variance, discrepancy masks) and (2) model-level analysis (SHAP attribution). Together, these components demonstrate both the discriminative power and the interpretability of defocus in distinguishing real from fake images.
    }
    \Description{
    An overview diagram illustrating the proposed defocus-based deepfake detection approach. 
    The top section shows the defocus map estimation process starting from a cropped face image, passing through edge detection, Gaussian gradient filtering at two different scales (\(\sigma = 1.5\) and \(\sigma = 2.0\)), and a guided filter module to produce a defocus blur map.
    The bottom section shows how this defocus map is used in downstream tasks such as deepfake detection, feature visualization using histograms and t-SNE, SHAP-based model explainability.
    }
    \label{fig:overview}
\end{figure*}
To characterize the defocus-based differences between real and fake images, we adopt three analysis methods: a binary difference mask to localize strong defocus discrepancies, local variance to quantify spatial consistency, and t-SNE~\cite{van2008visualizing} to visualize feature separability.

\paragraph{\textbf{Binary Mask Visualization of Defocus Discrepancy.}}
To provide a clear visual representation of the defocus differences between real and fake image pairs, we construct a binary mask that highlights regions where the absolute defocus value difference exceeds a threshold of 0.1. Specifically, we assign a value of 1 to pixels where the real and fake defocus maps differ by at least 0.1, and 0 elsewhere. This binary mask effectively localizes areas of significant defocus discrepancy, allowing intuitive identification of regions where synthetic images deviate from the physical defocus patterns observed in real images.

\paragraph{\textbf{Local Variance Analysis.}}\label{sec:local_variance}
To assess spatial consistency in defocus blur, we compute the local variance of the defocus map. In real images, defocus typically transitions smoothly, producing low variance across local regions. In contrast, manipulated or spliced regions in fake images often cause abrupt changes in defocus, leading to locally high variance. The resulting variance map thus serves as an effective indicator for detecting forgery-induced inconsistencies. This metric allows us to quantify subtle structural irregularities that are not easily captured by global statistics. Moreover, local variance analysis can highlight forged regions even when the overall defocus magnitude remains within a plausible range.

\paragraph{\textbf{t-SNE Visualization.}}  
We apply t-Distributed Stochastic Neighbor Embedding (t-SNE)~\cite{van2008visualizing} to the global features extracted from the defocus branch. The high-dimensional feature vectors are projected into a 2D space, allowing us to visually analyze the separability of real and fake images. This method helps identify whether the model can distinctly separate these classes based on the learned defocus features. The visualization also reveals the compactness of the learned feature clusters, providing an understanding of how the model organizes similar data points.

\subsection{Defocus Blur Map Model-level Analysis}
To interpret the model's decision-making process, we perform a \textbf{Model-level Interpretation} using SHAP~\cite{lundberg2017unified} for attribution-based analysis.

\paragraph{\textbf{Shapley Value (SHAP) Analysis.}}  
To analyze the contribution of individual features to the model's prediction, we employed SHapley Additive Explanations (SHAP) using the gradient explainer. This explainer is suitable for differentiable models and computes pixel-wise attributions based on gradients of the output with respect to the input. It enables saliency-based interpretation at the input level, offering fine-grained insight into the model’s decision process.

We applied the SHAP analysis to both the RGB-based and defocus-based models to allow modality-wise comparison of feature attribution. A total of 1,000 test samples were selected from each class (real and fake), and 10 real sampled images from the training set were used as the background distribution for SHAP estimation. Saliency maps were generated by overlaying the resulting SHAP values onto the corresponding input images. These maps serve as the basis for subsequent explainability analysis.

\paragraph{\textbf{Defocus–SHAP Alignment Analysis.}}
To evaluate how well the model’s saliency corresponds to physically meaningful defocus cues, we define two alignment metrics based on histogram comparisons between SHAP activations and pixel-wise defocus differences.

\begin{itemize}
    \item $D_{\text{real}}, D_{\text{fake}}$: pixel-wise defocus maps of real and fake images,
    \item $S_{\text{fake}}$: SHAP value map for the fake image (negatives clipped to zero),
    \item $M_{\text{diff}} = |D_{\text{fake}} - D_{\text{real}}|$: pixel-wise defocus difference map.
\end{itemize}

We define $N$ equally spaced bins over the defocus value range $[0, 1]$ as follows:
\begin{equation}
\text{bin}_i = \left[ \frac{i-1}{N}, \frac{i}{N} \right), \quad i = 1, \ldots, N
\label{eq:bin}
\end{equation}

Using these bins, we compute weighted histograms of $D_{\text{fake}}$ with $M_{\text{diff}}$ and $|S_{\text{fake}}|$ as weights, respectively. Specifically, $H_{\text{diff}}$ aggregates the defocus difference values $M_{\text{diff}}(x)$ across each bin and $H_{\text{shap}}$ accumulates the SHAP values $S_{\text{fake}}(x)$ in each bin as follows:
\begin{align}
H_{\text{diff}}[i] &= \sum_{x \in \text{bin}_i} M_{\text{diff}}(x) \label{eq:hdiff} \\
H_{\text{shap}}[i] &= \sum_{x \in \text{bin}_i} S_{\text{fake}}(x)
\label{eq:hshap}
\end{align}

For each bin, we sum the values of $M_{\text{diff}}(x)$ and $S_{\text{fake}}(x)$ for all locations $x$ whose defocus values fall within the bin. These histograms represent how real/fake distribution differences and model attribution intensities are distributed across the defocus spectrum. We denote the normalized histograms as $\hat{H}_{\text{diff}}$ and $\hat{H}_{\text{shap}}$, representing probability distributions over defocus bins. These are used to compute KL divergence and Alignment scores for evaluating how well model attributions align with distributional differences in defocus.
We define the overlap between the two distributions as follows:
\begin{equation}
\mathrm{Alignment} = \sum_i \min\left( \hat{H}_{\text{shap}}[i], \hat{H}_{\text{diff}}[i] \right).
\label{eq:align}
\end{equation}

\subsubsection{KL Divergence.}
We compute the Kullback–Leibler divergence between the normalized distributions.
The KL divergence is computed as follows:
\begin{equation}
\mathrm{KL}(\hat{H}_{\text{shap}} \parallel \hat{H}_{\text{diff}}) =
\sum_i\hat{H}_{\text{shap}}[i]\cdot\log\left(\frac{\hat{H}_{\text{shap}}[i]+\varepsilon}{\hat{H}_{\text{diff}}[i]+\varepsilon}\right)\label{eq:kl},
\end{equation}
where $\varepsilon$ is a small constant for numerical stability. These metrics quantify how closely the model’s attention (as expressed by SHAP) aligns with spatially meaningful defocus differences.

\section{Experiments}
\subsection{Dataset}
\paragraph{\textbf{FaceForensics++ (FF++)}} We use the \textit{raw} version of the FaceForensics++ (FF++)~\cite{rossler2019faceforensics++} dataset, which preserves high visual quality and includes subtle facial manipulations. Since only the face regions are modified while the background remains untouched, it is well-suited for evaluating local artifacts such as defocus blur mismatches. We used 1,000 real videos and an equal number of fake videos for each manipulation type (Deepfakes, Face2Face, FaceSwap, NeuralTextures) in a 1:1 real-to-fake ratio. All videos were converted into frames, and the dataset was split into training, validation, and test sets without any overlap between splits. All images are resized to 299×299 and normalized using ImageNet statistics.

\begin{table}[t]
\centering
\caption{Composition of real and synthetic images in the SYNDOF~\cite{lee2019deep} and LFDOF~\cite{ruan2021aifnet} datasets across train/validation/test splits.
Real images include CUHK~\cite{shi2014discriminative}, Flickr~\cite{lee2019deep}, Middlebury~\cite{scharstein2014high}, and LFDOF~\cite{ruan2021aifnet}, 
while synthetic images in SYNDOF are generated from SYNTHIA~\cite{ros2016synthia} and MPI~Sintel~\cite{butler2012mpi}.
Overall, the splits are constructed to maintain a nearly balanced ratio between real and synthetic samples.}
\small
\setlength{\tabcolsep}{3pt} 
\renewcommand{\arraystretch}{1.05}
\begin{tabular}{lcccccc}
\toprule
\multirow{2}{*}{\textbf{Split}} 
& \multicolumn{4}{c}{\textbf{Real}} 
& \multicolumn{1}{c}{\textbf{Synthetic}} 
& \multirow{2}{*}{\textbf{Total}} \\ 
\cmidrule(lr){2-5} \cmidrule(lr){6-6}
& CUHK & Flickr & LFDOF & Mid. & SYNDOF& \\
\midrule
Train & 416 & 1,928 & 4,676 & 185 & 7,224 & R: 7,205 / F: 7,224 \\
Val.  & 46  & 214   & 519   & 20  & 802   & R: 799 / F: 802     \\
Test  & 200 & ---   & 725   & 22 & 1,000 & R: 947 / F: 1,000   \\
\midrule
\rowcolor{gray!20} Total & 662 & 2,142 & 5,920 & 227 & 9,026 & R: 8,951 / F: 9,026 \\
\bottomrule
\end{tabular}
\footnotesize \\
\label{tab:dataset_splits_small}
\end{table}

\begin{figure}[t]
    \centering
    \includegraphics[width=1\linewidth]{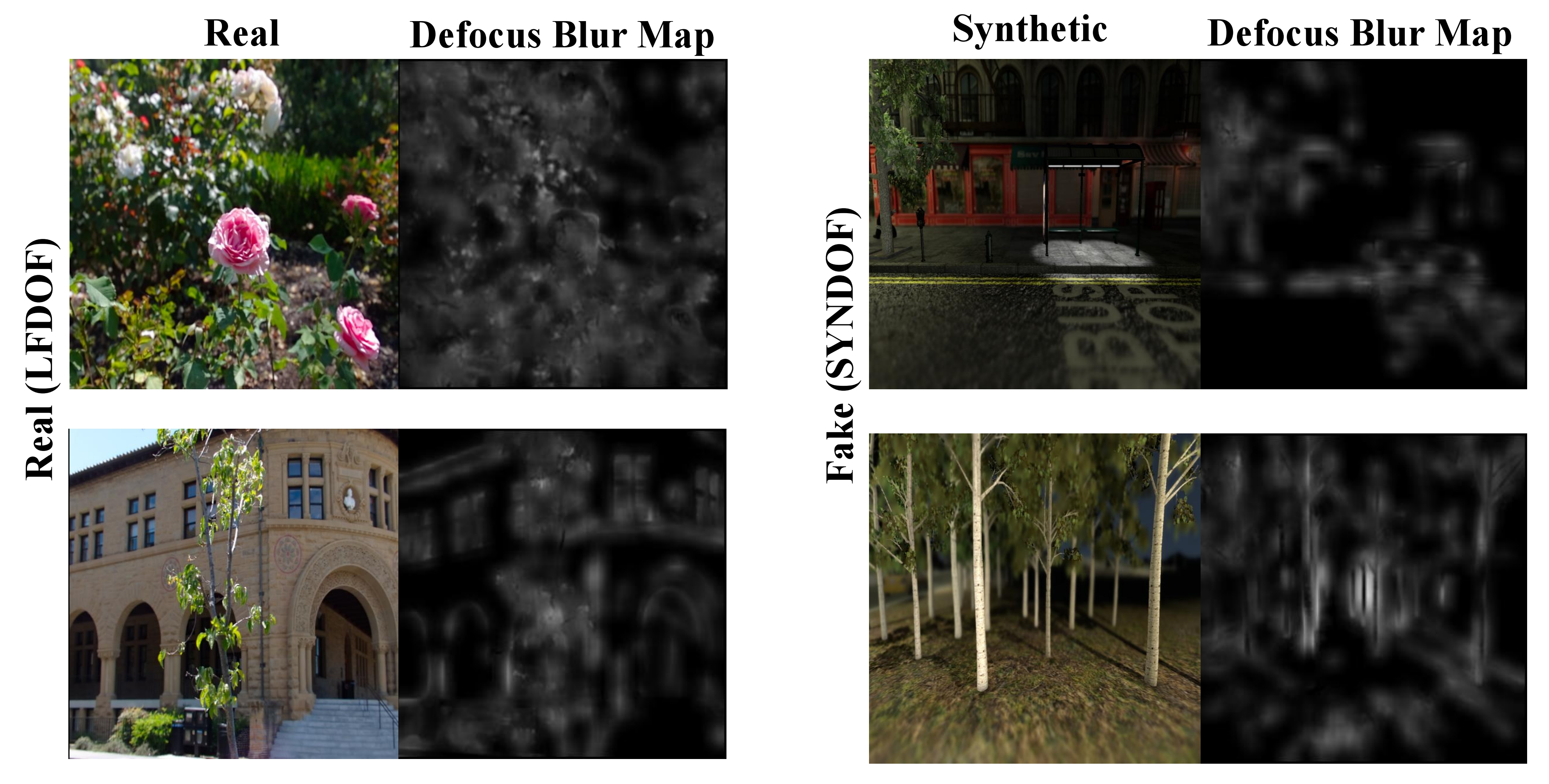}
    \caption{Comparison of Defocus blur maps from real and synthetic images. Real images show spatially distributed high-blur regions, whereas synthetic images lack realistic depth-dependent blur and appear globally sharper. Left: examples from the real-world LFDOF dataset. Right: examples from the synthetic SYNDOF (SYNTHIA~\cite{ros2016synthia}) dataset. In the defocus maps, brighter areas indicate stronger blur.
    }
    \Description{
    Comparison of Defocus Blur Maps from Real and Synthetic Images.
    }
    \label{fig:syndof}
\end{figure}

\paragraph{\textbf{SYNDOF}} The SYNDOF dataset~\cite{lee2019deep} was originally developed for training CNN-based defocus estimation models and contains both real and synthetic images.
For our experiments on binary classification of real and fake defocus blur, we reorganized the folder structure to suit our needs.
Middlebury~\cite{scharstein2014high} images, originally located under \texttt{test/SYNDOF} in the SYNDOF dataset release, were moved to \texttt{train/Middlebury} because they are real.
Since the number of real samples was insufficient, we supplemented them by sampling 5,920 images from LFDOF~\cite{ruan2021aifnet}, ensuring a balanced ratio of real to fake (1:1) across all splits.
LFDOF was included in both the train/validation and test splits. The aforementioned Middlebury data was used only in train/validation, and the RTF dataset was used exclusively for testing. (See Table ~\ref{tab:dataset_splits_small})
Real images are sourced from CUHK~\cite{shi2014discriminative}, Flickr~\cite{lee2019deep}, LFDOF, and Middlebury~\cite{scharstein2014high}, all of which naturally contain depth of field (DoF) blur resulting from optical capture conditions.
Synthetic images are generated by applying artificial defocus blur to sharp images from SYNTHIA~\cite{ros2016synthia} and MPI Sintel~\cite{butler2012mpi}.
As shown in Figure~\ref{fig:syndof}, synthetic images fail to reproduce realistic depth-dependent defocus, often appearing overly sharp. This discrepancy highlights the physically grounded difference between real and synthetic imagery and serves as a key basis for our defocus-based classification approach.

\subsection{Experimental Setup}
\paragraph{Deepfake Detection}
We evaluate defocus-only models on the FaceForensics++ dataset, using Xception as the primary backbone. Input images are resized to 299×299, and defocus blur maps generated via an optical cue–based module are fed into the classifier. Models are trained with a batch size of 32 using Adam optimizer and BCEWithLogitsLoss. We report the best model checkpoint selected by early stopping, which typically converged within 5 epochs. For comparison, we also test ResNet-50~\cite{he2016deep}, EfficientNet-B4~\cite{tan2019efficientnet}, and ViT-16~\cite{dosovitskiy2020image}, all under the same training settings.
\paragraph{AIGC Detection}
For the SYNDOF dataset, we use a dual-branch model combining RGB and defocus inputs. Each branch is based on ResNet-50, and feature maps are fused at the feature level before classification. Input sizes follow the native resolution of each backbone (299×299 for Xception, 224×224 for others). All models are trained with batch size 128, using Adam and BCEWithLogitsLoss.

\subsection{Detection Performance and Evaluation Metrics}
We evaluate the performance of our method using the following metric:
\begin{itemize}
    \item \textbf{Area Under the Curve (AUC):} A threshold-independent metric that measures the model's ability to distinguish between real and fake images. A higher AUC indicates better binary classification performance. AUC is used throughout both dataset evaluation.

    \item \textbf{Accuracy (Acc.):} The proportion of correctly classified samples over the total. Accuracy is used in the SYNDOF experiment to evaluate overall classification performance across synthetic manipulation types.

    \item \textbf{Recall:} The proportion of correctly detected fake samples, reflecting the model’s sensitivity to manipulated content. Recall is reported alongside accuracy in the SYNDOF experiment, where precision is nearly perfect and AUC may be unreliable.
\end{itemize}

\subsection{Detection Performance}
\subsubsection{\textbf{Deepfake Detection}}
\begin{table}[t]
\centering
\caption{Performance comparison on the FF++ (raw) dataset. The results of the compared methods are taken from DeepfakeBench~\cite{DeepfakeBench_YAN_NEURIPS2023}.
In the table, \textbf{N} represents Naive methods, \textbf{F} represents Frequency-based methods, and \textbf{S} represents Spatial-based methods.}
\resizebox{\columnwidth}{!}{%
\begin{tabular}{cllcccc}
\toprule 
\textbf{Type} & \textbf{Method} & \textbf{DF} & \textbf{F2F} & \textbf{FS} & \textbf{NT} & \textbf{Avg. AUC} \\
\midrule
N & Meso4~\cite{afchar2018mesonet} & 0.677 & 0.905 & 0.595 & 0.570 & 0.686 \\
N & MesoIncep~\cite{afchar2018mesonet} & 0.854 & 0.809 & 0.742 & 0.652 & 0.764 \\
N & CNN-Aug~\cite{wang2020cnn} & 0.905 & 0.879 & 0.903 & 0.731 & 0.854 \\
N & Xception~\cite{rossler2019faceforensics++} & 0.980 & 0.979 & 0.983 & 0.939 & 0.970 \\
N & EfficientB4~\cite{tan2019efficientnet} & 0.976 & 0.976 & 0.980 & 0.931 & 0.965 \\
S & Capsule~\cite{nguyen2019capsule} & 0.867 & 0.867 & 0.873 & 0.780 & 0.846 \\
S & FWA~\cite{li2018exposing} & 0.921 & 0.900 & 0.884 & 0.812 & 0.879 \\
S & X-ray~\cite{li2020face} & 0.793 & 0.987 & 0.987 & 0.929 & 0.924 \\
S & FFD~\cite{dang2020detection} & 0.824 & 0.978 & 0.985 & 0.931 & 0.929 \\
S & CORE~\cite{ni2022core} & 0.979 & 0.980 & 0.982 & 0.934 & 0.968 \\
S & RECCE~\cite{cao2022end} & 0.980 & 0.978 & 0.979 & 0.936 & 0.968 \\
S & UCF~\cite{yan2023ucf} & 0.988 & 0.984 & 0.990 & 0.944 & 0.976 \\
S & QAD~\cite{le2023quality} & 0.998 & 0.992 & 0.995 & \underline{0.984} & 0.992 \\
F & F3Net~\cite{qian2020thinking} & 0.979 & 0.980 & 0.984 & 0.935 & 0.969 \\
F & SPSL~\cite{liu2021spatial} & 0.978 & 0.975 & 0.983 & 0.930 & 0.966 \\
F & SRM~\cite{luo2021generalizing} & 0.973 & 0.970 & 0.974 & 0.930 & 0.961 \\
S+F & SFCF~\cite{zheng2025spatio} & \underline{0.998} & \textbf{0.999} & \textbf{0.999} & 0.983 & \underline{0.994} \\
\midrule
\rowcolor{gray!20}S & Defocus (Ours) & \textbf{0.999} & \underline{0.998} & \underline{0.998} & \textbf{0.999} & \textbf{0.998} \\
\bottomrule
\end{tabular}%
}
\label{tab:ff++_performance}
\end{table}

Table~\ref{tab:ff++_performance} presents a comprehensive performance comparison on the FaceForensics++ (raw) dataset, including various baseline and state-of-the-art detection methods. Our proposed method, denoted as \textbf{Defocus (Ours)}, achieves the highest average AUC score of \textbf{0.998}, outperforming both spatial-based and frequency-based approaches. Specifically, our model surpasses advanced hybrid models such as SFCF~\cite{zheng2025spatio} (0.994) and QAD~\cite{le2023quality} (0.992), while performing competitively across all manipulation types. 
The 95\% DeLong confidence intervals ~\cite{65f9f828-9f33-36dc-9429-5d215792ea89} for each manipulation type were extremely narrow: DF [0.9996–0.9997], F2F [0.9980–0.9987], FS [0.9981–0.9988], and NT [0.9989–0.9995].
Notably, our method achieves an AUC of 0.999 on the NT subset, which is considered the most challenging due to subtle manipulation artifacts. These results highlight the effectiveness of incorporating physically grounded defocus blur as a discriminative cue for deepfake detection.

\subsubsection{\textbf{AI-generated Content (AIGC) Detection}}
\begin{table}[t]
\centering
\caption{Impact of defocus feature integration on the SYNDOF dataset.
This table reports accuracy, recall, and AUC for RGB-only and RGB+Defocus models across three backbone architectures. Incorporating defocus features consistently improves performance across all backbones.
}
\label{tab:syndof_performance}
\resizebox{\columnwidth}{!}{%
\begin{tabular}{lccccc}
\toprule
\textbf{Method} & \multicolumn{2}{c}{\textbf{RGB}} & \multicolumn{3}{c}{\textbf{RGB + Defocus (Ours)}} \\
\cmidrule(lr){2-3} \cmidrule(lr){4-6}
& \textbf{Acc.} & \textbf{Recall} & \textbf{Acc.} & \textbf{Recall} & \textbf{AUC} \\
\midrule
Xception~\cite{chollet2017xception} & 0.951 & 0.906 & \cellcolor{gray!20}\textbf{0.972} (+2.1\%) & \cellcolor{gray!20}\textbf{0.946} (+4.0\%) & \cellcolor{gray!20}0.998\\
ResNet-50~\cite{he2016deep}         & \textbf{0.965} & \textbf{0.933} & 0.968 (+0.3\%) & 0.938 (+0.5\%) & \textbf{0.999}\\
ViT-16~\cite{dosovitskiy2020image}  & 0.736 & 0.489  & 0.758 (+2.2\%) & 0.536 (+4.7\%) & 0.953 \\
\bottomrule
\end{tabular}
}
\end{table}

To evaluate generalization beyond facial forgeries, we conducted experiments on the SYNDOF dataset, a fully synthetic benchmark with defocus effects applied through post-processing. We compared RGB-only models with RGB+Defocus multi-branch architectures across three backbones (Xception, ResNet-50, ViT-16). As shown in Table~\ref{tab:syndof_performance}, incorporating defocus blur consistently improved accuracy and recall across all backbones. 
Xception achieved the highest absolute performance (Acc. 0.972, Recall 0.946), while ViT-16 exhibited the largest relative gains (+2.2\% Acc., +4.7\% Recall). 
These results highlight that defocus blur can serve as an effective auxiliary module, providing complementary physical cues to RGB features for robust AIGC detection.

\subsubsection{\textbf{Ablation Study}}
\paragraph{\textbf{Ablation Study on Feature Extractors}}
\begin{table}[t]
\centering
\caption{Ablation study on backbone architectures for defocus-based detection on the FaceForensics++ dataset.
AUC results across four manipulation types (DF, F2F, FS, NT) show that defocus features are effective with all backbones, with the Xception model achieving the highest performance under identical training settings.
}
\resizebox{\columnwidth}{!}{%
\begin{tabular}{ccccccc}
\toprule 
\textbf{Type} & \textbf{Backbone} & \textbf{DF} & \textbf{F2F} & \textbf{FS} & \textbf{NT} & \textbf{Avg. AUC} \\
\midrule
\rowcolor{gray!20}S & Defocus (Xception) & \textbf{0.999} & \textbf{0.997} & \textbf{0.996} & \textbf{0.998} & \textbf{0.9975} \\
S & Defocus (Efficient-B4) & 0.999 & 0.997 & 0.995 & 0.997 & 0.9970 \\
S & Defocus (ViT-16) & 0.986 & 0.975 & 0.970 & 0.964 & 0.9737 \\
\bottomrule
\end{tabular}%
}
\label{tab:Backbone_ablation}
\end{table}
To assess the robustness of our defocus-based detection method across different backbone architectures, we conducted an ablation study on the FaceForensics++ (FF++) dataset.
Specifically, we compared the performance of three widely used feature extractors: Xception, EfficientNet-B4, and ViT-16.
The results, summarized in Table~\ref{tab:Backbone_ablation}, show that while all models benefit from defocus-based inputs, Xception achieves the highest AUC across all manipulation types.
EfficientNet-B4 also performs competitively, whereas ViT-16 lags behind, possibly due to its weaker inductive bias in modeling local blur patterns.
These findings suggest that defocus features are generally effective, but their utility can be further enhanced by using backbones optimized for capturing local image statistics.

\paragraph{\textbf{Gaussian Filter Tuning for Detection Performance}}
To analyze the effect of the Gaussian filter on detection performance, we conducted experiments with various combinations of Gaussian kernel standard deviations ($\sigma_1$, $\sigma_2$) in the Gaussian gradient-based Defocus Blur Map generation process.
Smaller $\sigma$ values preserve fine edge details but may introduce noise, while larger values overly smooth edges and degrade informative signals.
To isolate the impact of these parameters, we performed the ablation based on the Xception backbone, which showed the best performance in Table~\ref{tab:syndof_performance}.
Experimental results showed that increasing $\sigma$ significantly reduced performance (e.g., accuracy and recall dropped to 0.965 and 0.933, respectively), while the baseline setting ($\sigma_1 = 1.5$, $\sigma_2 = 2.0$) achieved the best balance between accuracy (0.972) and recall (0.946).
On the other hand, using smaller filter values such as $(\sigma_1 = 1.0, \sigma_2 = 1.5)$ led to under-smoothing, resulting in degraded recall (0.911) despite high AUC (0.999), likely due to increased sensitivity to noise. These results are summarized in Table~\ref{tab:gaussian_ablation}.
Since minimizing false negatives is crucial in deepfake detection, recall was prioritized in selecting the final configuration.
\begin{table}[t]
\centering
\caption{Ablation study results on Gaussian filter $\sigma$ selection for the Xception model. The (1.5, 2.0) configuration achieves the best trade-off between accuracy and recall.}
\resizebox{\columnwidth}{!}{%
\begin{tabular}{lcccc}
\hline
\textbf{Model} & \textbf{Filter Parameters ($\sigma_1$, $\sigma_2$)} & \textbf{Accuracy} & \textbf{Recall} & \textbf{AUC} \\
\hline
\multirow{3}{*}{Xception} 
& (1.0, 1.5) & 0.954 & 0.911 & \textbf{0.999} \\
&\cellcolor{gray!20} \textbf{(1.5, 2.0)} & \cellcolor{gray!20}\textbf{0.972} & \cellcolor{gray!20}\textbf{0.946} & \cellcolor{gray!20}0.998 \\
& (2.0, 2.5) & 0.965 & 0.933 & \textbf{0.999} \\
\hline
\end{tabular}
}
\label{tab:gaussian_ablation}
\end{table}

\subsection{Defocus Blur Map Feature-level Analysis}
To evaluate whether defocus blur patterns differ between real and manipulated images, as posed in \textbf{RQ1}, we conduct three feature-level experiments: binary mask analysis, local variance measurement, and t-SNE visualization. These experiments are designed to quantitatively and visually assess the separability between real and fake images in the defocus blur space.

\subsubsection{\textbf{Defocus Value Difference Binary Mask}}
\begin{figure}[t]
    \centering
    \includegraphics[width=\linewidth]{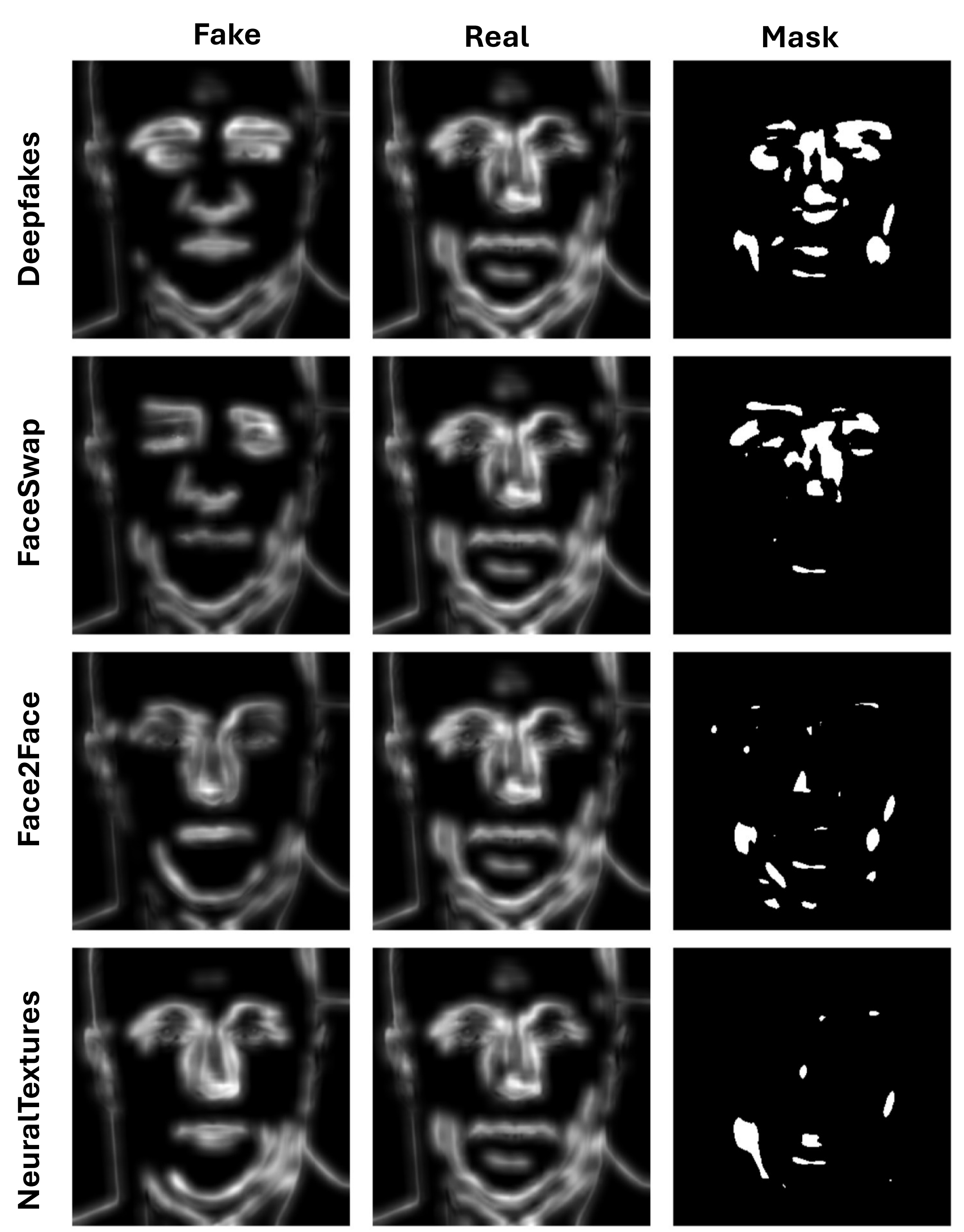}
    \caption{Binary mask visualization of defocus differences between real and fake images. The masks highlight regions where synthetic manipulations introduce unnatural or inconsistent blur, providing a clear visual cue for forgery detection. Pixels are marked as 1 if the absolute defocus value difference between real and fake maps exceeds 0.1.}
    \Description{
    To intuitively highlight regions with substantial defocus discrepancies between real and fake images, we computed a pixel-wise absolute difference map between the normalized real and fake defocus blur maps. A threshold of 0.1 was applied, where all pixels showing a difference greater than this threshold were marked as 1 in the binary mask, indicating significant defocus change regions. This visualization allows us to localize areas where synthetic manipulations introduce unnatural or inconsistent defocus patterns, which can serve as potential forensic cues for deepfake detection.on of the SHAP-related patterns is provided in a later section.
    }
    \label{fig:binary_mask}
\end{figure}
The binary masks in Figure~\ref{fig:binary_mask} clearly demonstrate that defocus changes exist between real and fake images, with manipulated samples exhibiting regions of unnatural or inconsistent blur. To quantify and analyze these differences, we examined binary defocus masks derived from manipulated images across four manipulation methods—DF, FS, F2F, and NT—each compared against a common real reference image. Each manipulation type includes 6,400 fake images, which are compared pixel-wise with their corresponding real counterparts.

We varied the threshold from 0.1 to 0.0001 to measure the average number of activated pixels. The results show that DF and FS exhibit localized and concentrated discrepancies at higher thresholds, whereas F2F and NT show a sharp increase in activated regions at lower thresholds, indicating subtle and spatially distributed inconsistencies. This suggests that, although these methods may appear less invasive at a coarse level, they in fact induce fine-grained and spatially dispersed defocus changes, likely due to localized expression modifications. For visualization, the binary masks in Figure~\ref{fig:binary_mask} were generated using a threshold of 0.1 to highlight differences discernible to the human eye.

\subsubsection{\textbf{Local Defocus Variance Analysis}}
\begin{figure}[t]
    \centering
    \includegraphics[width=\linewidth]{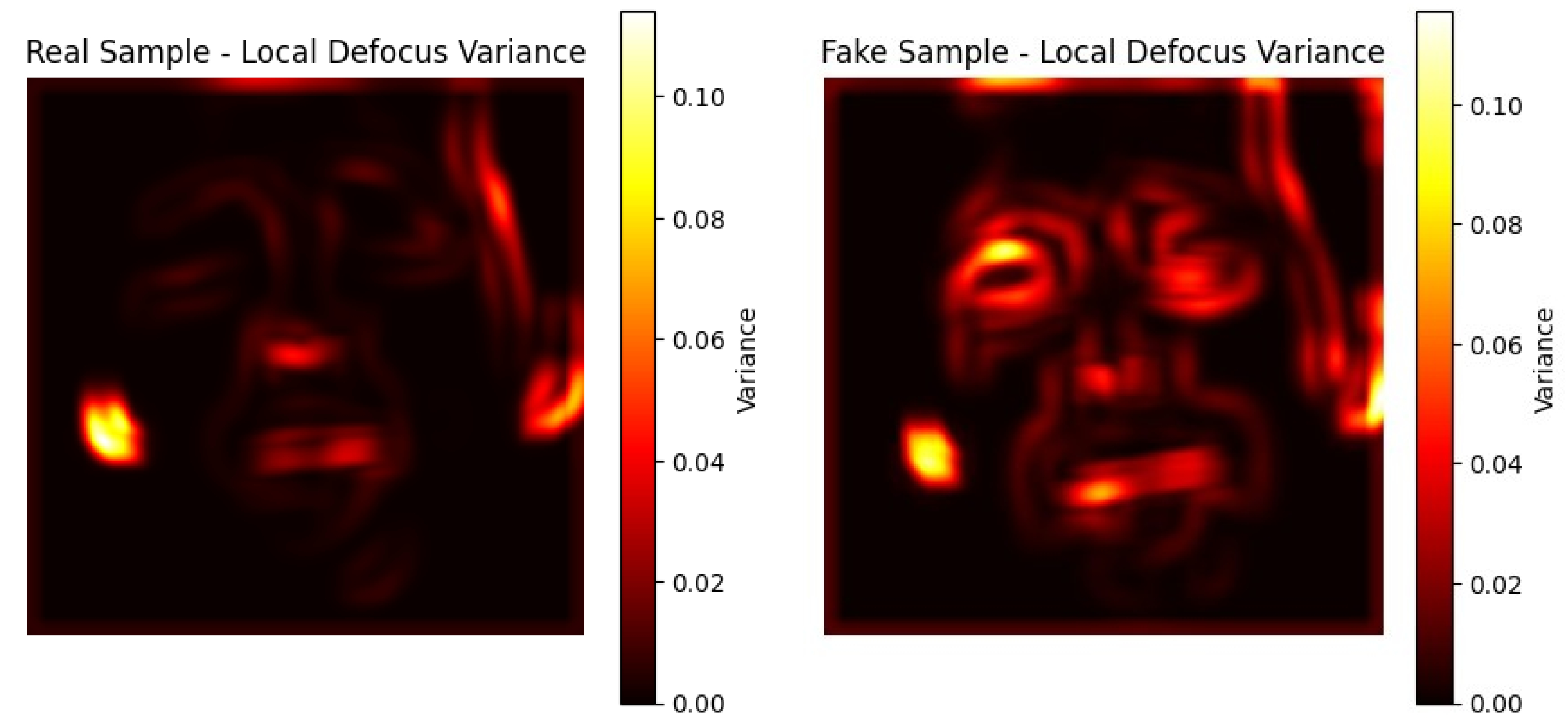}
    \caption{Local variance heatmaps of real and fake images. Fake images exhibit higher local variance with abrupt and inconsistent spikes, whereas real images show smooth and uniformly low variance along natural facial boundaries.}
    \label{fig:defocus_variance}
\end{figure}

Fake defocus maps exhibit significantly higher local variance than real ones, reflecting unnatural transitions in blur intensity. Figure~\ref{fig:defocus_variance} illustrates an example of a local variance map, where real images show smooth and uniformly low variance while fake images contain abrupt spikes.

To further validate this observation, we quantitatively evaluated its discriminative power using a two-sample Kolmogorov–Smirnov (KS) test. The test was conducted on 12,800 images (6,400 real and 6,400 fake), using the average local variance computed per image. The test yielded a KS statistic of $D = 0.0492$ and a $p$-value of $3.68 \times 10^{-7}$, indicating a statistically significant difference between the two distributions. These results support the hypothesis that local variance patterns extracted from defocus maps are effective cues for distinguishing between real and fake images.

\subsubsection{\textbf{Global Statistical Analysis (Real vs Fake)}}
\begin{figure}[t]
    \centering
    \includegraphics[width=0.7\linewidth, height=50mm]{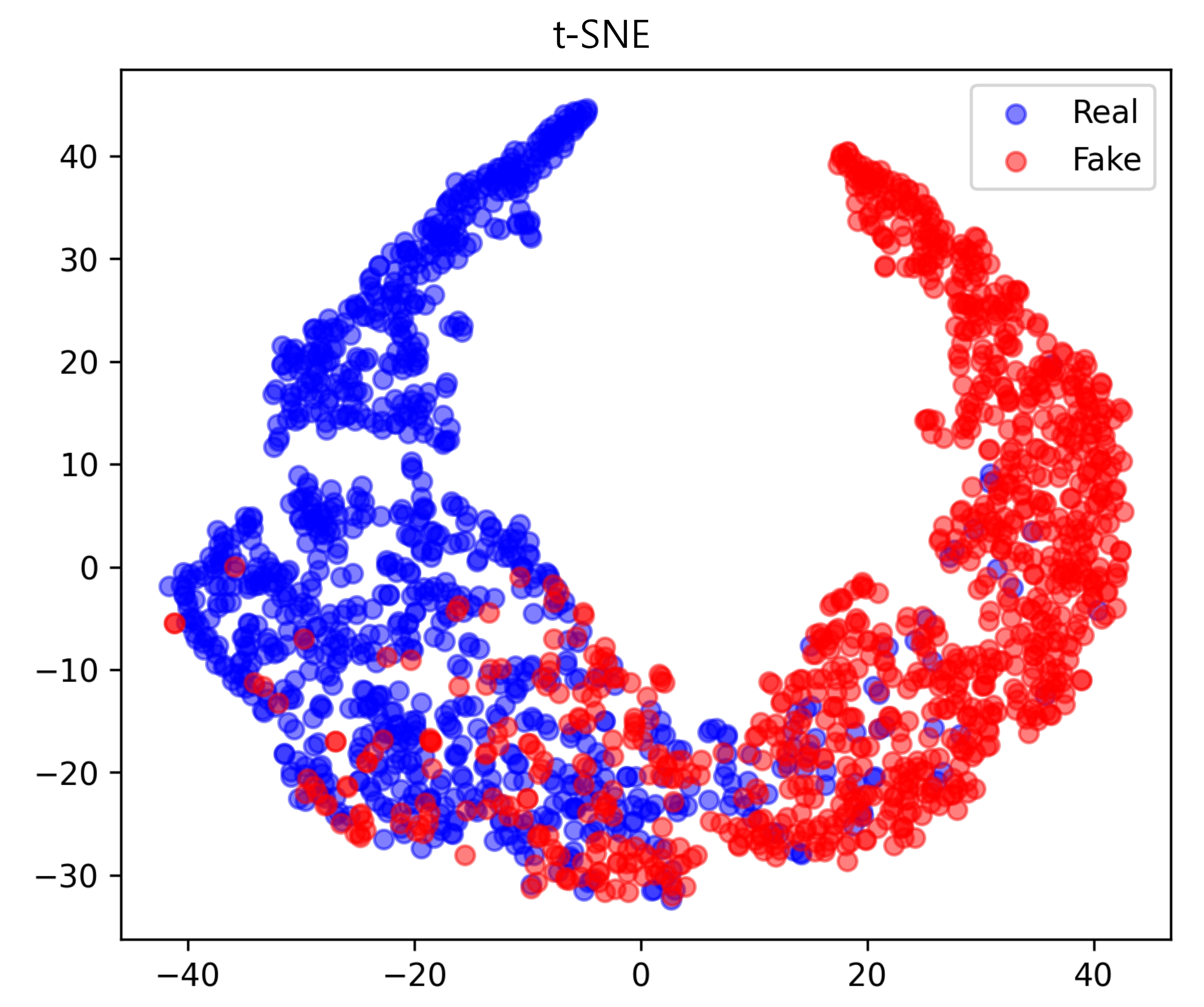}
    \caption{t-SNE visualization of defocus blur features from real and fake images. Real and fake samples form clearly separated clusters, demonstrating that defocus blur characteristics provide discriminative information for deepfake detection. This visualization is obtained by applying t-SNE to features extracted from 1,000 real and 1,000 fake images.}    \label{fig:defocus_tsne}
\end{figure}
As shown in Figure~\ref{fig:defocus_tsne}, real and fake images form clearly separated clusters based on their defocus blur features. This separation demonstrates that defocus characteristics carry discriminative information useful for distinguishing authentic from manipulated content. To obtain this result, we applied t-SNE to features extracted from 1,000 real and 1,000 fake images sampled from the test set. These findings indicate that defocus blur may serve as a meaningful cue for deepfake detection and holds potential as a complementary feature in automated forensic systems.

\subsection{Model-level Analysis}
To address \textbf{RQ3}, which investigates whether the trained model actively utilizes defocus-based cues in its decision-making process, we conduct a model-level analysis focusing on interpretability and attribution methods. Specifically, we employ SHAP value analysis and visual explanations to examine how defocus-related features influence the model's predictions.
\subsubsection{\textbf{Shapley Value Visualization}}

\begin{figure}[t]
    \centering
    \includegraphics[width=1\linewidth]{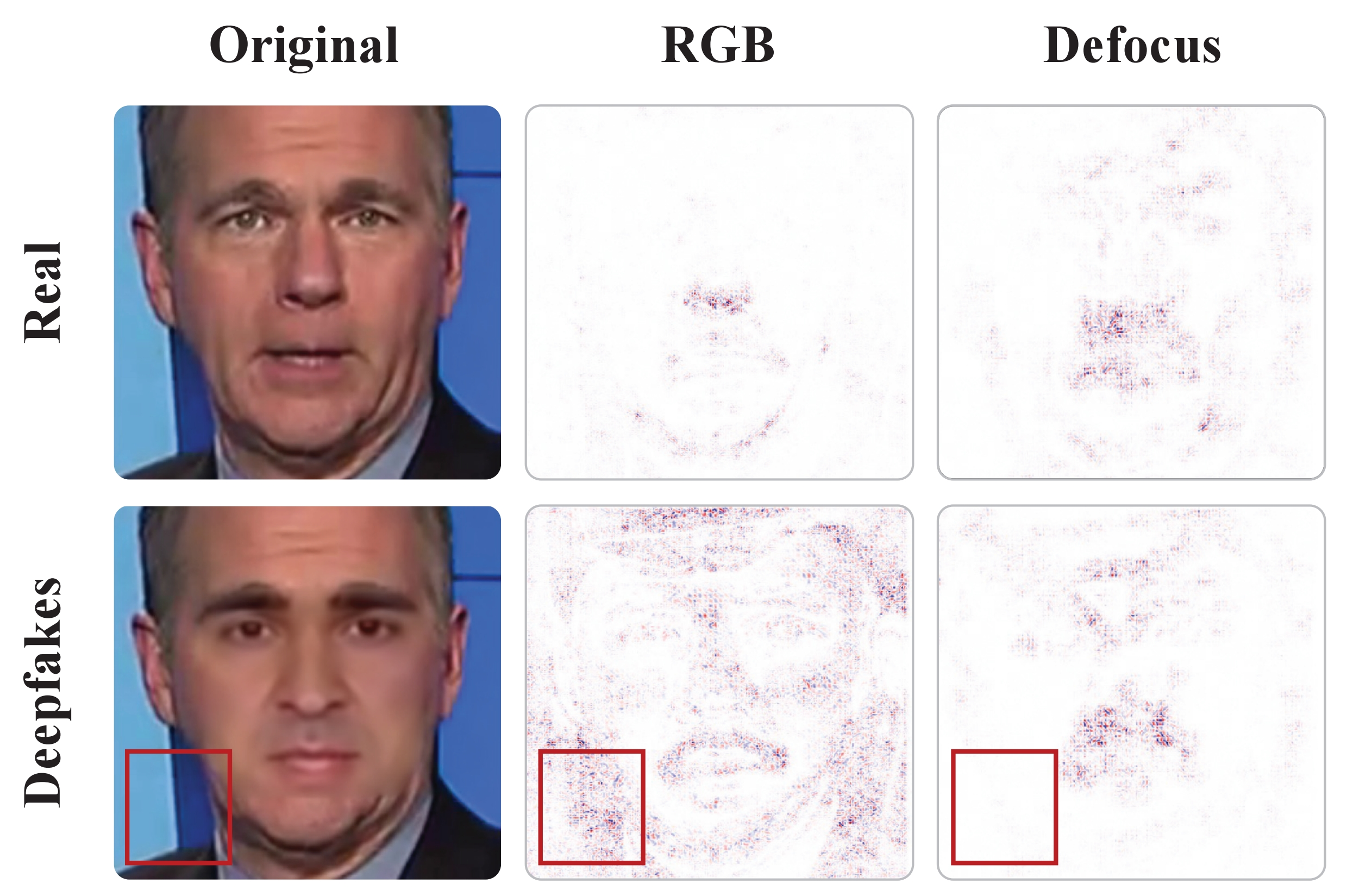}
    \caption{SHAP attribution visualization of RGB and Defocus models.
    The Defocus model highlights manipulated facial regions, whereas the RGB model attends to background areas, demonstrating the superior interpretability and forensic relevance of defocus-based detection.}
    \label{fig:shapley_visualization}
\end{figure}

As shown in Figure ~\ref{fig:shapley_visualization}, the SHAP analysis reveals that the Defocus-based model concentrates on meaningful discrepancies in the manipulated face region, whereas the RGB-based model mainly relies on unaltered background areas. In other words, the Defocus-based model leverages physically grounded and manipulation-related cues, such as facial contours and blur transitions. This allows it to perform more interpretable and reliable classification, while the RGB-based model often fails to capture the critical features of manipulation by focusing on overall image structure or background. This contrast highlights the superiority of the Defocus-based model in terms of interpretability and generalizability, as it actively attends to physically plausible forensic signals.

\subsubsection{\textbf{SHAP Values Alignment Score}}\label{sec:shap_kde_alignment}
\begin{figure}[t]
    \centering
    \includegraphics[width=\linewidth]{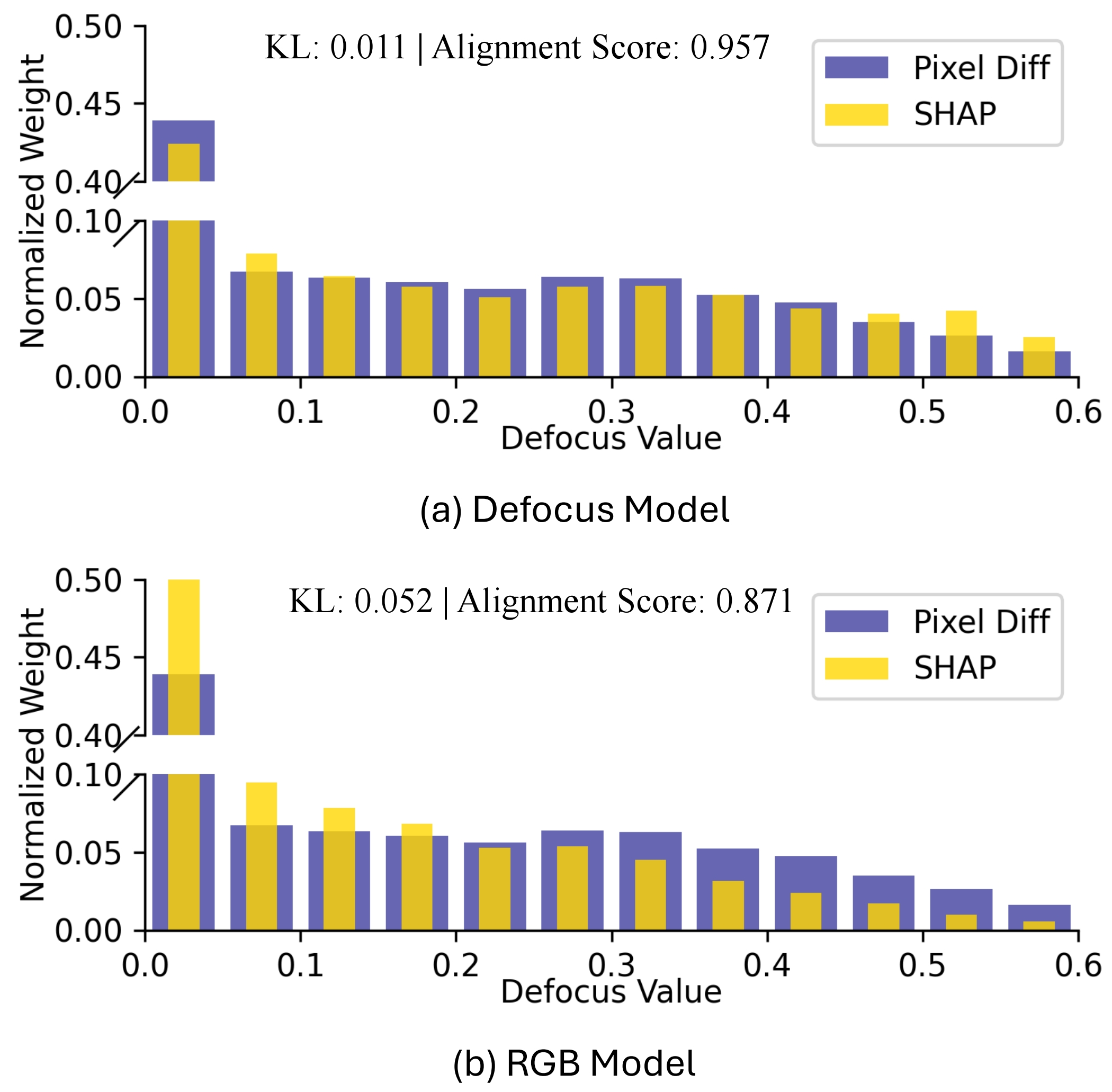}
    \caption{Alignment between defocus differences and SHAP attributions in fake images. The Defocus model exhibits stronger alignment with physically meaningful blur transitions than the RGB model, indicating its ability to capture unrealistic defocus patterns in synthetic content. Purple bars denote defocus differences, and yellow bars indicate SHAP activations.}
    \label{fig:shapley_alignment}
\end{figure}
The defocus-based model exhibited a stronger alignment with physically meaningful blur transitions compared to the RGB-based model, achieving a higher alignment score (0.822 vs. 0.679) and a lower KL divergence (0.188 vs. 0.307). This indicates that the model more effectively attends to unrealistic blur patterns commonly observed in synthetic images. To quantitatively validate this effect, we compared SHAP-based saliency maps with pixel-wise defocus differences using 1,000 samples from each manipulation type (DF, F2F, FS, NT). As shown in Figure~\ref{fig:shapley_alignment}, the distribution of SHAP activations (yellow) exhibits a high degree of correspondence with the distribution of defocus differences (purple), confirming that the model attends to physically meaningful blur discrepancies. This alignment is further substantiated by the high alignment score.

\section{Limitation and Future Work}
This study focuses on the discovery of defocus blur as a new modality for deepfake detection. Its robustness in practical deployment settings remains important direction for future work. We consider two major real-world conditions:  \textbf{(1) compression environments} such as severe JPEG compression and  \textbf{(2) capture environments} such as sensor noise, motion blur, illumination changes, and lens distortion. In compression environments, aggressive JPEG compression on social media can cause blocking and ringing artifacts that hinder defocus blur estimation. In our C40 evaluation, the method achieved competitive accuracy but fell short of state-of-the-art performance, indicating that compression artifacts interfere with edge-based defocus blur estimation. In capture environments, motion blur from camera shake, noise from low-light or high-ISO settings, and optical distortions from wide-angle or low-quality lenses can distort or obscure genuine defocus patterns. These artifacts are rarely reproduced in synthetic data because of their lack of aesthetic necessity. Future work will incorporate realistic post-capture artifacts from real photographs into training and evaluation. We will also develop robust defocus blur estimation techniques and distortion-aware augmentation strategies.

\section{Conclusion}
Our work demonstrates the effectiveness of defocus blur as a physically grounded and interpretable feature for deepfake detection. We propose a defocus map representation based on optical principles and validated its utility through feature- and model-level analyses addressing three research questions. Feature-level experiments using Binary masks, variance maps, and t-SNE revealed consistent differences between real and fake images. Model-level SHAP analyses showed that the defocus-based model attends to regions with significant blur discrepancies, confirmed by low KL divergence and high overlap scores. Our method achieved state-of-the-art performance (AUC 0.998) on FF++ with defocus-only input, and further improved performance on the SYNDOF dataset using an RGB+Defocus multi-branch architecture, demonstrating its potential as a complementary modality to existing approaches. These findings establish defocus blur as a generalizable and interpretable signal for trustworthy image forensics. To ensure reproducibility, we released the full evaluation code and results.

\begin{acks}
This work was partly supported by Institute for Information \& communication Technology Planning \& evaluation (IITP) grants funded by the Korean government MSIT:
(RS-2022-II221199, RS-2022-II220688, RS-2019-II190421, RS-2023-00230337, RS-2024-00356-293, RS-2024-00437849, RS-2021-II212068,  RS-2025-02304983, RS-2024-00436936, and  RS-2025-02263841).
\end{acks}

\clearpage
\section*{GenAI Usage Disclosure}

\textbf{Text Rephrasing Assistance:}
The authors used ChatGPT (GPT-4o) to assist in rephrasing technical summaries into fluent academic English. All outputs were carefully reviewed, edited, and validated by the authors to ensure accuracy and clarity.

\textbf{Image Assistance:}
Portions of Figure~1 were generated with the aid of ChatGPT (GPT-4o). Specifically, the bookshelf and white door elements in the bottom-right background were created based on text prompts and incorporated into the final illustration.

\textbf{Equation Formatting:}
Select equations in Section~3 were partially derived using ChatGPT (GPT-4o). The tool was provided with mathematical descriptions and code fragments and returned LaTeX-style expressions, which were subsequently verified and refined by the authors.

\textbf{Code Assistance:}
The authors also used ChatGPT (GPT-4o) to assist in writing and debugging Python scripts used for SHAP analysis, statistical evaluation, and visualization. All generated code was thoroughly validated, modified as needed, and integrated by the authors into the final experimental pipeline.

\bibliographystyle{ACM-Reference-Format}
\balance
\bibliography{sample-base}


\end{document}